# Face Detection and Face Recognition In the Wild Using Off-the-Shelf Freely Available Components

Hira Ahmad

*Abstract*— This paper presents an easy and efficient face detection and face recognition approach using free software components from the internet. Face detection and face recognition problems have wide applications in home and office security. Therefore this work will helpful for those searching for a free face off-the-shelf face detection system. Using this system, faces can be detected in uncontrolled environments. In the detection phase, every individual face is detected and in the recognition phase the detected faces are compared with the faces in a given data set and recognized.

*Index Terms*— face detection, face recognition, free software,

## I. INTRODUCTION

Face detection and face recognition is an interesting area of research. These systems have advantages over all the manual person authentication systems and in fact, have advantage on many of the automated systems. These systems range from image processing and computer vision. Face recognition is important because this technology provides safety and security while detecting and identifying persons. Face recognition requires less processing than many other automated systems like eye scanner, finger reader and other applications. The accuracy rate in face recognition technolo- gies are far more than before. The goal of computer vision researchers are to make automated systems that can surpass human accuracy on detecting and recognizing human faces. Humans can detect all faces very easily but recognize only those faces, which they have already seen and can tag or label the face corresponding with the name in their brains. Similarly, the automated systems can detect all the faces and recognize only those which are already in their databases, and can label or tag the new arrival test face image only if this image is already available in the database. The comparison of the faces will return a match. Human faces have wide range of diversity such that even twins are distinguishable by a human naked eye. The humans, sometimes, cannot recognize faces at a very first glance. It is difficult for an automated system to detect and recognize the faces so precisely, however, the recent researchers are putting an effort to improve the precision and accuracy rates of such systems.

We are proposing the algorithm for face detection and face recognition. Face detection is to locate the boundary of a human face, thus finding all the faces in an image. Face recognition is to identify the faces from the given database and label them correctly. An image recognition systems, particularly facial recognition systems require that the processing images should have high quality. However, some factors are always present like bad lightings, poor illumination, complex background, size variations, pose variance, orientations and occlusions etc., which badly impact the face detection and recognition.

## II. RELATED WORK

Face detection is the basic step for face recognition systems. The researchers of the past and the present have done immense work and contributed alot in both these areas. General object detection and recognition techniques [9] [19] [39] [8] can also be used for face detection and recognition with some modifications, while all the methods used are actually system dependent. Face detection and face recognition are specific techniques while object detection and recognition are general techniques.

### A. Face Detection

Different techniques for face detection have been introduced so far, and the specific technique is used for different scenarios. Many state of the art techniques and methods were mentioned [11] for face detection. Jesorsky et al. [13] proposed a shape comparison approach, their system consisted of two phases, a coarse detection and a refinement phase, where each of them is consisted of a segmentation and a localization step. Masi et al. [23] had done face recognition on huge dataset with millions of images, they claimed that large training dataset is necessary for face recognition. Zhu et al. proposed a single model [42] for face detection, pose estimation and landmark localization. Their model was based on a blend of tree-structured part models. They modelled every facial landmark as one part and used global mixtures to record topological changes. Ranjan et al. [26] presented a single deep convolutional neural networks (CNN) for simultaneous face detection, landmarks localization, pose estimation and gender recognition. In their proposed network, the features in CNN were distributed hierarchically. Jiang et al. [14] proposed to apply the Faster RCNN. They trained a Faster R-CNN model on WIDER face dataset, and reported the state of the art results. Viola et al. [40] had done three contributions. They used an integral image, image representation technique, which helped to compute features quickly.They used AdaBoost learning algorithm to select a small number of features from a large set of features. Lastly, they combined classifiers to form a cascade which discarded background regions quickly and did computation on face like regions. Lienhart et al. used the extended set

This work was supported by ITU Graduate Fellowship Grant.

H Ahmad is graduate student in Information Technology University, Lahore, Pakistan. mscs18006@itu.edu.pk

of Haar features and showed the 10% low false alarm rates [17]. The authors [5], analyzed the performance of public domain classifiers of Open CV and presented that different classifiers are better for different situations. They presented heuristics to increase the facial detection rate. Erdem et al. [7] combined Haar detector with skin filter for face detection, which showed good results. Dawoud et al. [6] presented a fast template matching technique based on Optimized Sum of Absolute Difference method for face localization. Li et al. presented a CNN cascade architecture [16] which worked at various resolutions, rapidly rejected the background regions in low resolution stages and evaluated challenging candidates on high resolution stages. Sun et al. [29] proposed a three level convolutional network for the estimation of the face keypoints. Zhang et al. [41] proposed tasks-constrained deep model method which had optimized face detection and over- come the pose variation and occlusions problems. Ranjan et al. [25] proposed a deep pyramid deformable part model which is used to detect faces of various sizes and poses in unconstrained conditions. Shaban et al. [27] presented a multi- person head segmentation technique which they used a convolutional encoder-decoder, their network was trained to assign high probability to head pixels and low probability to non-head pixels. The real time face detection and tracking algorithm [2] was implemented by using Arduino board, Web cam and Microcontroller devices which was then aimed to develop a real time application. Another framework used, [30] which was introduced for detecting, localizing, and classifying faces in terms of visual traits, e.g., sex or age, from arbitrary viewpoints and in the presence of occlusion.

B. Face Recognition

Apart from face detection, many researchers have contributed and gave their new methods for face recognition. Soltanpour et al. [28] surveyed on the local feature extraction methods for the three dimensional face recognition, they considered features of the face very important for the recognition of faces. Much research has been done till now on face recognition. Cao at al. [4] presented a Deep Residual EquivAriant Mapping block, that had added residuals to the input representation for the transformation a profile facial representation to the canonical pose for face recognition purpose. Sparse kernel learning algorithm [33] was presented to automatic the selection and integration of the most discriminative subset of kernels which were derived from different image set representations. Periocular regions, [37] [38] [31] [3] the regions around the eyes based person identification in videos were also considered as the image classification particularly face classification problem. Parkhi et al. [24]. used deep CNN with proper training achieved state of the art results of face recognition. Multi- Order Statistical Descriptors [22] were used for real time face recognition and object classification. Multi-pose aware CNN deep learning models [1] were used to generate multiple pose specific features, pose specific CNN features were assembled for the purpose of facial recognition. Hierarchical sparse spectral clustering method, [20] a dimensionality reduction approach, was used for image set classification. Haxby et al. [10] proposed a model for the organization of the system that emphasizes a distinction between the representation of invariant and changeable aspects of faces. The Gabor-Fisher classifier [18] method achieved 100 % accuracy on face recognition using only 62 features, employed an enhanced Fisher discrimination model on an augmented Gabor feature vector, which was derived from the Gabor wavelet transformation of face images. Hu et al. [12] presented sparse approximated nearest points (SANP) and generated intermediate sample representations by using a sparse set of samples from the sets. Set classification was performed by simple nearest neighbour based approaches using the SANPs. Semi-supervised Spectral Clustering [21] technique was proposed for the classification of Image Set. Li et al. [15] presented pore scale facial features , method which despite of the high variations in expression and pose, outperformed a number of face recognition method. Uzair et al. [34] used three dimensional cosine transform for features extraction from hyper spectral data and partial least squares techniques for face recognition. Uzair et al. [35] presented image sets representation that was used for object classification, the proposed representation was applied to various biometric applications. Hyperspectral face recognition algorithm [32] was proposed which used spatiospectral covariance and partial least square classification. Uzia et al. [36] described that the spectral reflectance, which is some measurement of wavelength of the electromagnetic energy, was not a reliable biometric for face recognition in unconstrained conditions.

III. PROPOSED SYSTEM

Most of the face recognition packages used in our algorithm are taken from this website https://face-recognition.readthedocs. io/en/latest/face_recognition.html# module-face_recognition.api. Our proposed system takes the coding related help from this website https://github.com/ageitgey/face_ recognition. Till now, many new techniques and methods have been implemented and modified by the developers and researchers, but sometimes it gets very challenging to use which method for the specific purpose, we have introduced a very simple and efficient method for face detection and face recognition.

Our proposed system consists of mainly three steps: (i) Finding location of faces (ii) Encoding (iii) Comparison of Faces

A. Finding location of faces

Face detection is to find the locations of the person from the input query images, we have used the Haar cascade classifier for the face detection purpose. The Haar Cascade Classifier contains series of stages where each stage comprises of the series of weak learners. At the first, weak learners will always detect all the faces but can also detect the non facial images. All the detected facial images with some non facial images were given to the next classifier, the next classifier is the strong classifier from the previous one, which also detect faces, but rejects some of the non facial images which were considered as faces in the previous classifier. The Haar Cascade classifier is to detect the faces accurately, and its primary purpose is to detect the faces and rejects all the false positive non facial images. These classifiers combined to form a cascase architecture. Below is the diagram of the Haar Cascade Classifier working.

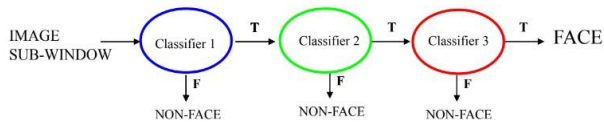

Figure I: Haar Cascade Classifier

B. Encoding

Our data set contains thirty individual images. The individual faces in the data set are labelled with the names of the respective person. We have not compared directly that faces in our algorithm, rather we have encoded first all the faces. The features, while encoding are extracted from the high resolution individual images, and these are the selected features which helps in the recognition of faces. These individual faces are encoded by the face encoding module and after the encoding process, we have saved the new data set of these 30 encoded images. The encoding function converts the image to 128 dimensional byte vector for all the examples present in the data set, however more dimensions can be selected and may lead to more accuracy, but it also requires extensive training, 128 dimensions byte vector is the ideal for the system. The characteristics of these encoded images are : (i) All the images contain 128 key features which should be compared with the new arrival detected location of input test images in our algorithm. (ii) The size of each encoded image in the the data set is of two Kilo bytes.

C. Comparison of Faces

The encoded images are get compared with the input query image, all the detected faces in the query images are get compared by the known encoded faces, at one time, and the particular face location is get labelled by the respective names. The comparison has been done using the Euclidean Distance formula, we have applied the threshold while comparing the faces, if the euclidean distance is smaller or equal to some threshold we have set, we will declare a match, otherwise it will be a non match. If there is a match, the output image showed the bounded box on that face with the respective label behind that face, if there is not a match, then the detected faces in the input image will get compared to the next database image.If all the database images are compared, the next detected face of the input image will be compared to all the encoded database images. The output image will show the bounded box over all the identified images with their specified labels. The threshold, we have set is a very small value, keeping the fact that smaller the euclidean distance, higher the matching.

The mathematical formula for calculating the Euclidean distance is:

$$d(a, b) = \sqrt{(bx - ax)^2 + (by - ay)^2} \quad (1)$$

where d represents the Euclidean distance and (x, y), (a, b) are the two coordinates. The general idea is, if the comparing images are exactly same, the calculated euclidean distance will be zero. The minimum the euclidean distance, the more the two comparing images are similar.

IV. DATA SET

Our database consists of thirty facial images. The images in the databse are of students and are taken under controlled environments. We have not directly compared these images, rather first encoded these images into the minimal size of two kilobytes per image. These encoded images are then compared to the encoded version of the comparing facial image.

Below are the high resolution images and in the JPEG format, which gets encoded after while for comparison.

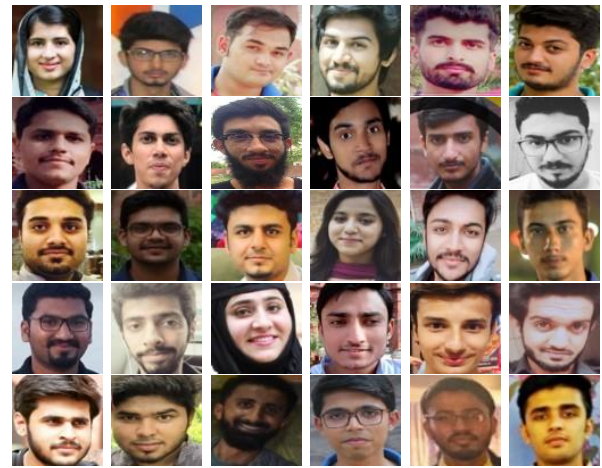

Figure II: Data Set

The accuracy of the face detection and face recognition module of our algorithm mainly depends on the face orientation in the images, the features or characteristics will be lost if the face orientation is not frontal. The frontal view of the faces are more desirable for our algorithm, otherwise the results can be different. The images in the data set should have these characteristics: (i) The images should be captured in the feasible environment, so that the images don't have blurriness, occlusions, sharp changes, background clutter etc., (ii) The face or faces in the images should have a frontal view (iii) The images should be captured where light is appropriate (iv) The images must be of high resolution. If these data characteristics will not be fulfilled, we will face terrible problems in either detecting or recognizing the faces.

## V. EXPERIMENTS AND RESULTS

The faces that gets compared with the data set are labelled and recognized. The eight input test images are given to our system one by one, and it shows satisfactory results of face detection and face recognition. The accuracy on seventh test image is less due to some problems of occlusions and all faces are not having the frontal view, one face is detected twice in the sixth input image, besides this the other images have shown some good accuracy results of face detection and face recognition.

We have done testing on the eight test images. These test images, called group photos, represented as GP. We have implemented our algorithm in the Python language with Open CV library. Below are the results, when our algorithm is implemented on the eight tested images.

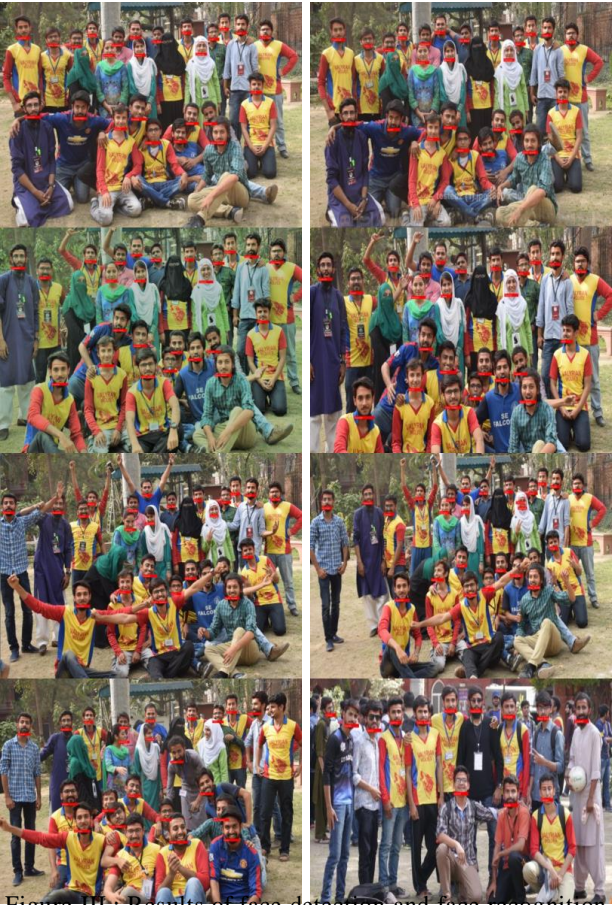

Figure III : Results of face detection and face recognition on eight test images

## VI. ACCURACY ON PROPOSED SYSTEM

A. Accuracy of Face Detection

The face detection accuracy is calculated on the eight input test images, which we called as group photos. We have counted manually the total faces for every test image. The total faces in the group photos are written in the Total faces column, and the detected faces are written in the column Detected faces. The face detection accuracy, we define is, (detected faces/total faces) x 100. FP, TP, FN and TN columns in the data set represents false positive, true positive, false negative and true negative respectively. There is no false positive and false negative problems in these sets of images, rather group photo fulfills the true positive and true negative conditions. False positive, means that there is a background in the image, but detected as a face. This condition does not hold in any of the group photos, which we want. True positive, states that the face is detected as a face. In our case, this condition holds, as only faces are detected as a face, no problem with that too. False negative depicts that face is considered as a background. This condition also does not hold in any of the test images or group photos as desired. True negative means background is detected as background, and this holds true, as desired. The overall accuracy of face detection is the average of all the accuracy's. For our case, the accuracy of face detection is 97.5%. The group photos, named as 5 and 7 has shown 92 % accuracy, the group photo
6 has shown 96% accuracy, while the remaining group photos have shown 100 % accuracy. The results can be improved if we test on more test images, but for know, we have tested only on eight test images.

TABLE I: Face Detection Accuracy

| GP | FP | TP | FN | TN | Total faces | Detected faces | Accuracy |
|---|---|---|---|---|---|---|---|
| 1 | × | D | × | D | 25 | 25 | 100% |
| 2 | × | D | × | D | 25 | 25 | 100% |
| 3 | × | D | × | D | 25 | 25 | 100% |
| 4 | × | D | × | D | 25 | 25 | 100% |
| 5 | × | D | × | D | 25 | 23 | 92% |
| 6 | × | D | × | D | 25 | 24 | 96% |
| 7 | × | D | × | D | 25 | 23 | 92% |
| 8 | × | D | × | D | 11 | 11 | 100% |

B. Accuracy of Face Recognition

The test images contain faces of individuals/students which are present in my data set. The data set is actually the encoded faces of the individuals represented as C. C refers to the whole class individual images, described above. Total faces represent the faces, in the particular group photo which are going to be compared with the individual encoded photos. Similarly, $a_{pp}$ are the recognized faces from the total faces, $a_{aa}$ refers to the absent students, $a_{ps}$ refers to the present-stranger wrong mapping. It can happen that the student is a stranger to my data set and the algorithm label him/her with the present student, $a_{pp}$ refers to the absent-present student wrong mapping. It can happen that the student is present and the algorithm label him/her with the unknown or absent and $a_{as}$ refers to the absent-stranger wrong mapping. It can happen that the individual is a stranger to my data set and the algorithm label him/her with the absent student.These all wrong mappings affect the accuracy directly or indirectly. We define accuracy as the sum of $a_{pp}$ and $a_{aa}$ divided by C. For our case, the accuracy of face recognition part is 92.29%. Besides the group photo 6 and 7, all the group photos have accuracy more than 90%.

TABLE II: Face Recognition Accuracy

| GP | C | Total Faces | $a_{pp}$ | $a_{aa}$ | $a_{ps}$ | $a_{ap}$ | $a_{as}$ | Accuracy |
|---|---|---|---|---|---|---|---|---|
| 1 | 30 | 25 | 24 | 5 | 0 | 1 | 0 | 96.67% |
| 2 | 30 | 25 | 24 | 5 | 0 | 1 | 0 | 96.67% |
| 3 | 30 | 25 | 25 | 5 | 0 | 0 | 0 | 100% |
| 4 | 30 | 25 | 24 | 5 | 0 | 1 | 0 | 96.67% |
| 5 | 30 | 25 | 21 | 5 | 0 | 4 | 0 | 86.67% |
| 6 | 30 | 25 | 20.5 | 5 | 0 | 1 | 0 | 85% |
| 7 | 30 | 25 | 18 | 5 | 0 | 2 | 0 | 76.67% |
| 8 | 30 | 11 | 11 | 19 | 1 | 0 | 1 | 100% |

## VII. CONCLUSIONS

It is a fact that the choice of face detection and face recognition method depends on the application. The paper brings the algorithm that has minimized the computation time and increased the performance by comparing the test images by the encoded images of the size of only two kilobytes. The face detection accuracy of this algorithm is 97.5% and the face recognition method has accuracy of 92.29%. These accuracies are totally dependent on how we defined our metrics for the accuracy measurement described in the face detection and face recognition accuracy section.